\pdfoutput=1

\documentclass[11pt]{article}

\usepackage[]{ACL2023}

\usepackage{times}
\usepackage{latexsym}

\usepackage[T1]{fontenc}

\usepackage[utf8]{inputenc}

\usepackage{microtype}

\usepackage{inconsolata}
\usepackage{graphicx}
\usepackage[export]{adjustbox}
\usepackage{multirow}
\usepackage{booktabs}

\renewcommand{\thefootnote}{$\ast$}

\title{Label Alignment and Reassignment with Generalist Large Language Model for Enhanced Cross-Domain Named Entity Recognition}
\author{Ke Bao\textsuperscript{1}\footnotemark[1]\\
  School of Computer Science \\
  Fudan University \\
  \\\And
  Chonghuan Yang\textsuperscript{2}\\
  School of Computer Science \\
  Fudan University \\}

\begin{document}
\maketitle
\footnotetext[1]{Corresponding author}

\begin{abstract}
Named entity recognition on the in-domain supervised and few-shot settings have been extensively discussed in the NLP community and made significant progress. However, cross-domain NER, a more common task in practical scenarios, still poses a challenge for most NER methods. Previous research efforts in that area primarily focus on knowledge transfer such as correlate label information from source to target domains but few works pay attention to the problem of label conflict. In this study, we introduce a label alignment and reassignment approach, namely LAR, to address this issue for enhanced cross-domain named entity recognition, which includes two core procedures: label alignment between source and target domains and label reassignment for type inference. The process of label reassignment can significantly be enhanced by integrating with an advanced large-scale language model such as ChatGPT. We conduct an extensive range of experiments on NER datasets involving both supervised and zero-shot scenarios. Empirical experimental results demonstrate the validation of our method with remarkable performance under the supervised and zero-shot out-of-domain settings compared to SOTA methods.
\end{abstract}

\section{Introduction}
Named entity recognition (NER) aims to identify mentions of rigid designators within text that correspond to predefined entity types such as person, location, and organization, etc. \cite{NER-2007}, influencing a variety of downstream tasks in natural language processing \cite{NER4TU-2019, NER4Query-2011, NER4Summary-1999, NER4QA-2006, NER4MT-2003}. In recent years, deep learning and pretraining language models have garnered significant attention due to its success across various fields. A number of advanced NER methods based on deep models perform excellently \cite{W2NER-2021, InferNER-2021, PIQN-2022, Diffusionner-2023, promptNER-2023}, while a substantial amount of annotated data are available. However, these well-trained deep models usually struggle to perform as well on out-of-domain datasets.

\begin{figure}
    \includegraphics[width=0.5\textwidth, right]{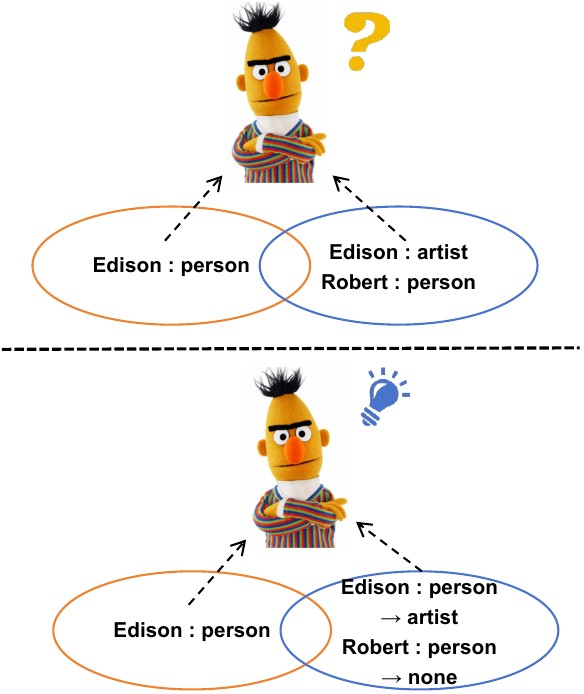}
    \caption{Due to the label conflict problem, the training without label alignment encounters conflicting knowledge among entities with similar context semantics on both the source and target domain datasets, while training with aligned labels can avoid this issue.}
    \label{fig:LabelConflict}
\end{figure}

To deal with these issues, some research works delve into cross-domain NER, which aims to study how to improve the performance on target domain with typically scarce supervised data by leveraging resource-rich source domain data. A portion of previous research works connect cross-domain NER and transfer learning, enabling deep networks to simultaneously adapt to data distributions of both source and target domains through multi-task learning or continual learning approaches \cite{Zero-2020, MDGA-2020, CrossDomainLM-2019, MLSTM-2020}. Some studies attempt to compare and correlate label information across source and target domain datasets to transfer type knowledge \cite{LANER-2022, GraphNER-2022, Cp-NER-2023, PLTR-2023}, while other studies get domain-specific entity knowledge through data augmentation techniques \cite{DA4NER2021, DAPT-2021, FactMix-2022}.

In this study, we continue to explore cross-domain NER from the label perspective, but we address the label conflict problem neglected by previous research, depicted in Figure \ref{fig:LabelConflict}. Target domain datasets typically have more Fine-grained annotation than that of source domain datasets, which leads to inconsistent entity knowledge. To deal with this problem, we introduce a label alignment and reassignment approach solving the knowledge conflict problem for enhanced cross-domain NER, the workflow of which is described in Figure 1. In the process of label alignment, our pretrained language model trained well on source domain distribution will predict pseudo labels with original types on target domain distribution. Then, we classify every entity type of target domain dataset into corresponding original entity types. After that, we obtain a relationship mapping table between source and target types. In the process of label reassignment, we make the model learn the mapping based on target domain datasets. In this way, we observe performance improvements across all target domain datasets, but there is still a problem that our model is not able to assign types with high confidence. Because it's a little difficult for it to discriminate between similar entity types (e.g., "person" and "politician" , "event" and "election") with limited supervised data. Fortunately, this problem can be effectively alleviated by incorporating an external knowledge source such as using a generalist LLM. In our experiments, we use GPT-3.5 to enhance the reassignment process that enables our method to perform in the zero-shot setting and observe substantial improvements in the supervised setting. We summarize the main contributions as follows:\\
1) We introduce a label alignment and reassignment approach for enhanced cross-domain named entity recognition, which addresses the label conflict problem and can be applied in both supervised and zero-shot scenarios.\\
2) We enhance type inference by leveraging GPT-3.5 in the process of label reassignment, enabling our method to perform excellently in the zero-shot setting.\\
3) Empirical experiments under supervised and zero-shot out-of-domain settings as well as the supervised in-domain setting show good performances, validating the effectiveness of our method.

\section{Related Work}
\subsection{Named Entity Recognition}
The studies on Named Entity Recognition can be categorized based on modeling paradigm. The most common approach is sequence labeling, which assigns a predefined tag (e.g., from the BIO scheme) to each token in a sequence \cite{SL-2015, SL-2007, SL-2020, Split-NER-2023}. Some studies reformulate NER as a span-level classification task, which involves enumerating all possible spans within a text and classifying them as valid mentions along with their types \cite{SB-2017, SB-2019, SB-2020, PL-Marker-2022}. Additionally, a number of studies \cite{HG-2015, HG-2016, HG-2018, HG2-2018}, design hypergraph structures to address overlapped NER by representing potential mentions in an exponential manner. Others utilize sequence-to-sequence generation paradigm modeling NER tasks \cite{SEQ-2015, SEQ-2019, SEQ-2020, SEQ-2021, SEQ-2022, ASP-2022}. Recently, unified information extraction frameworks have sparked heated debate and perform excellently on NER datasets \cite{UIE-2022, USM-2023}.

\begin{figure*}
    \includegraphics[width=\textwidth, right]{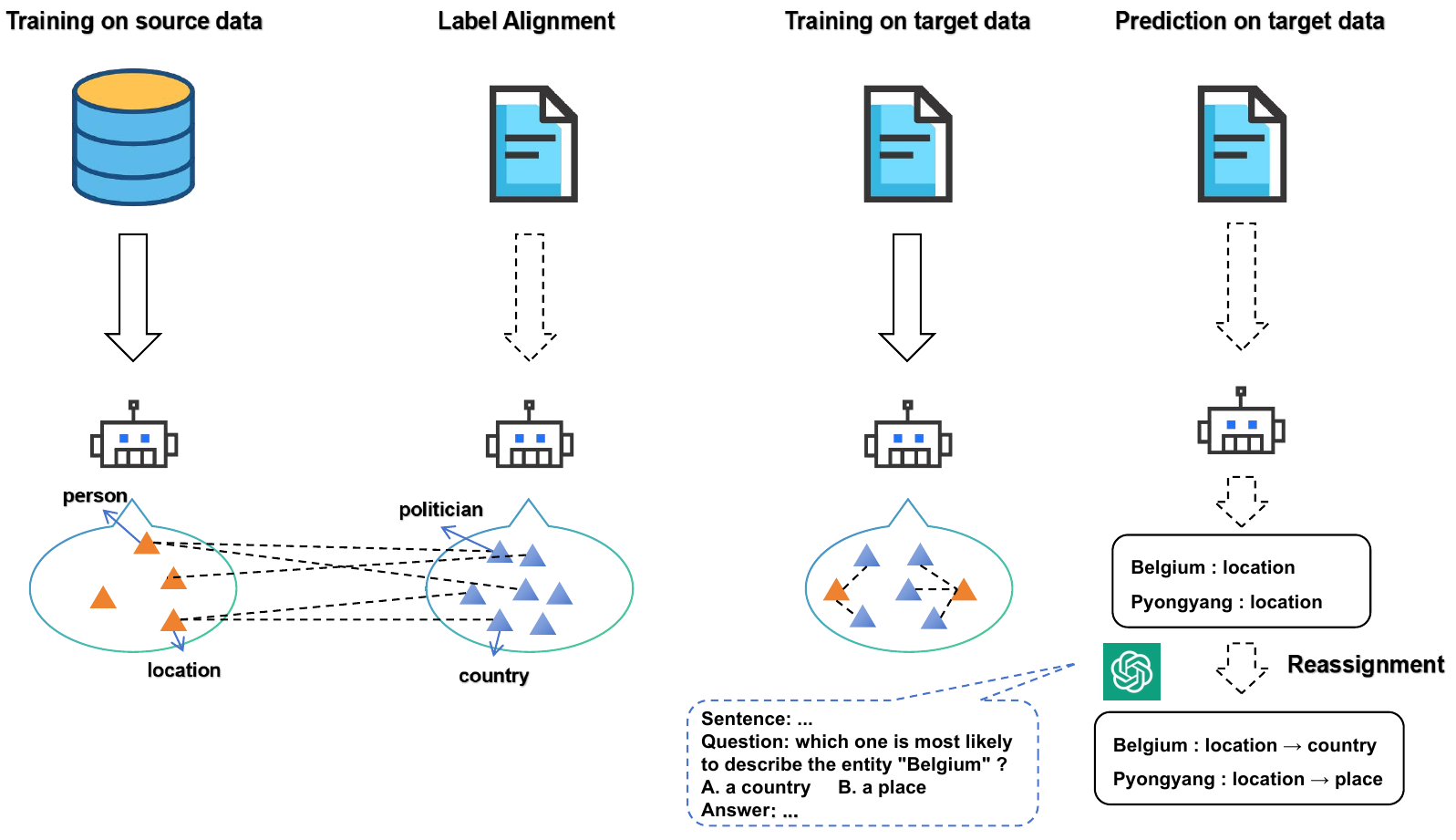}
    \caption{The workflow of our method, which includes all the steps from training to prediction. The solid arrows represent training processes, and the dotted arrows represent inference processes.}
    \label{fig:workflow}
\end{figure*}

\subsection{Cross-Domain Named Entity Recognition}
Domain-specific named entity recognition with limited supervised data is a more common scenario in practice. Therefore, cross-domain NER aiming to enhance the performance in specific-domain by leveraging rich domain-agnostic data, has drawn substantial attention in recent years. \citet{DAPT-2021} performs continuous training on a large unlabeled corpus related to target domains, which effectively improves model performance and can be applied to other methods. \citet{CrossDomainLM-2019} utilizes a parameter generation network to merge cross-domain language modeling with NER. \citet{LANER-2022} introduces an autoregressive paradigm to strengthen the connection between label and text. \citet{Cp-NER-2023} proposes a unified model for all target domain datasets based on collaborative domain-prefix tuning. \citet{FactMix-2022} uses rationale-centric data augmentation to improve the model's generalization ability. \citet{PLTR-2023}  automatically extracts entity type-related features to enhance knowledge transfer based on mutual information criteria.

\subsection{Using LLM on Named Entity Recognition}
Large-scale language models, e.g., ChatGPT, Gemini and LLaMA have demonstrated impressive performances in many fields due to their surprising emergent abilities \cite{GPT-4, Gemini, LLaMA-2023, EmergentAbility-2022}. \citet{ChatGPT4IE-2023} evaluates ChatGPT on a series of IE tasks. \citet{UniversalNER-2023} enables LLaMA-7B to generalize on unseen domains with the help of knowledge distillation from ChatGPT. \citet{GoLLIE-2023} provides annotation guides for types on zero-shot information extraction.

\section{Methodology}
\subsection{Preliminary}
Conventional NER tasks aim to extract an entity collection $E=\{e_1, ..., e_n\}$ from a given sentence $S=\{w_1, ..., w_m\}$. On this basis, the cross-domain NER task assumes that there are different data distributions, a well-sampled source dataset $\mathcal{D}_s$ and several low-resource target datasets $\mathcal{D}_t$. Typically, source domain data has more general entity types while target domain data is developed from a specific application field. Therefore, Cross-domain NER aims to improve the model performance on $\mathcal{D}_t$ by transferring entity knowledge from $\mathcal{D}_s$.

There are many prevailing extraction paradigms for named entity recognition. To keep up with the current development of language models and make it have good type extensibility, we explore the theory of label alignment and reassignment based on a sequence-to-sequence generation paradigm. Specifically, we use the identical architecture to ES-SDE introduced by \citet{SDE-2024} but without handling of unmentioned targets, which consists of an instruction $I$, an input sentence $S$, a textual label $Y$ with linear format of multiple predictions. We fine-tune a autoregressive pretrained language model to maximize the conditional probability $\mathcal{P}(Y|I, S)$.

\subsection{Label Alignment and Reassignment}
The overall workflow of our method is demonstrated in Figure \ref{fig:workflow}, which consists of four steps from training to prediction.

\textbf{Training on source domain.} As with other cross-domain NER methods, we first train a generative language model $\mathcal{M}$ on the source domain data $\mathcal{D}_s$ under the standard supervised setting. Every input sentence is added with the same simple instruction prefix which includes all the types on $\mathcal{D}_s$. After that, $\mathcal{M}$ is equipped with entity knowledge from $\mathcal{D}_s$. We name the initial model as $\mathcal{M}_s$.

\textbf{Label Alignment.} The step is to match every target entity type with the most relevant source entity type. Specifically, we use $\mathcal{M}_s$ to predict all samples in the train set of $\mathcal{D}_t$, obtaining corresponding pseudo labels. Then we compare pseudo labels and gold labels, and determine the mapping relationship between source domain types and target domain types based on the statistical results. For example, the type "musician" on $\mathcal{D}_t$ is assigned to the type "person" on $\mathcal{D}_s$ due to the fact that the majority of entities occurring in pseudo labels with type "musician" is predicted to be type "person" by $\mathcal{M}_s$. However, there are some entities with certain target entity types that may not be recognized at all on account of the limitation of source domain knowledge. To address this issue, we generalize these unrecognized types as miscellaneous. After this process, each target entity type should have one and only one corresponding source entity type. In the usual case, source entity types are more general than target entity types, resulting in a one-two-more mapping relationship. Additionally, we consider a zero-shot out-of-domain scenario without any labeled data on target domains. In this scenario, we can prompt a general large language model like ChatGPT to produce synthetic labels as the replacement of gold labels.

\textbf{Training on target domain.} We regard every sample on $\mathcal{D}_t$ as a triplet $(S, Y_s, Y_t)$, wherein $Y_s$ and $Y_t$ denote entity types on source domain and target domain respectively. In order to maximize the use of the prior knowledge of model $\mathcal{M}_s$, we modify the textualized labels only slightly, represented as $Y_s : S_{i:j} => Y_t$ called label reassignment. In this way, we perform supervised fine-tuning of $\mathcal{M}_s$ on the target domain dataset $\mathcal{D}_t$ to get a new model with target domain entity knowledge, namely $\mathcal{M}_t$.

\textbf{Prediction on target domain.} $\mathcal{M}_t$ has a basic target entity knowledge and is able to perform inference on unseen data. Theoretically, we can write the inference process with label reassignment as the dot product of $\mathcal{P_M}(Y_s, S_{i:j}|S, I)$ and $\mathcal{P_M}(Y_t|Y_s, S_{i:j}, S, I)$, which measure the model recognition degree of entity knowledge in source domain and target domain respectively. In practice, we observe that the values of the latter are usually less than that of the former on gold labels of target development sets. This means $\mathcal{M}_t$ has greater ability to recognize entities with types in source domain types than target domain. This phenomenon is easy to explain and understand. On the one hand, target entity types are more refined and complex than source entity types. On the other hand, target domain datasets have only a few supervised samples, on which fine-tuned language models get inadequately trained.

\subsection{Enhanced Inference with LLM}
Based on the previous assumption, if the discrimination ability between similar types could be effectively improved, the model performance will also be improved. Naturally, an external knowledge source will be helpful for this issue, such as resorting to API-based LLMs. Considering the conditional probability $\mathcal{P_M}(Y_t|Y_s, S_{i:j}, S, I)$, we can regard it as a classification task, the type collection of which depends on the mapping relationship. Furthermore, we transfer this classification task into a multiple-choice question that is very easy to understand by LLMs. A specific prompt example of the multiple-choice question is demonstrated in Figure \ref{fig:workflow}. What's more, this enhanced inference with LLM enables our method to adapt to unseen data distributions, greatly enriching its application scenario.

\begin{table}
\centering
\resizebox{0.48\textwidth}{!}{
\begin{tabular}{l|c|c|c|c}
\toprule
\multirow{2}*{\textbf{Dataset}} & \multirow{2}*{\textbf{Type Num}} & \multicolumn{3}{c}{\textbf{Sentence Num}}\\
 & & \textbf{Train} & \textbf{Dev} & \textbf{Test}\\
\midrule
\textbf{CoNLL03} & 4 & 14,987 & 3,466 & 3,684\\
\midrule
\textbf{Politics} & 9 & 200 & 541 & 651\\
\midrule
\textbf{Science} & 17 & 200 & 450 & 543\\
\midrule
\textbf{Music} & 13 & 100 & 380 & 456\\
\midrule
\textbf{Literature} & 12 & 100 & 400 & 416\\
\midrule
\textbf{AI} & 14 & 100 & 350 & 431\\
\midrule
\textbf{Movie} & 14 & 7,816 & - & 1,953\\
\midrule
\textbf{Restaurant} & 8 & 7,660 & - & 1,521\\
\bottomrule
\end{tabular}
}
\caption{\label{table-statistics}
The statistics of all datasets used in our experiments.
}
\end{table}

\begin{table*}
\centering
\resizebox{\textwidth}{!}{
\begin{tabular}{lcccccc}
\toprule
 \textbf{Model} & \textbf{Politics} & \textbf{Science} & \textbf{Music} & \textbf{Literature} & \textbf{AI} & \textbf{Average}\\
\midrule
FLAIR \cite{FLAIR-2018} & 69.54 & 64.71 & 65.60 & 61.35 & 52.48 & 62.73\\
Cross-domain LM \cite{CrossDomainLM-2019} & 68.44 & 64.31 & 63.56 & 59.59 & 53.70 & 61.92\\
DAPT \cite{DAPT-2021} & 72.05 & 68.78 & 75.71 & 69.04 & 62.56 & 69.63\\
LightNER \cite{LightNER-2021} & 72.78 & 66.74 & 72.28 & 65.17 & 35.82 & 62.56\\
LANER \cite{LANER-2022} & 71.65 & 69.29 & 73.07 & 67.98 & 61.72 & 68.74\\
Cp-NER \cite{Cp-NER-2023} & 73.41 & \textbf{74.65} & \textbf{78.08} & \underline{70.84} & 64.53 & \underline{72.30}\\
UniNER-7B \cite{UniversalNER-2023} & 66.90 & \underline{70.80} & 70.60 & 64.90 & 62.90 & 67.22\\
\midrule
LAR-base w/o EI & 73.80 & 68.82 & 74.30 & 69.46 & 63.85 & 70.05\\
LAR-base & \underline{78.34} & 66.27 & 74.40 & 69.74 & \underline{66.35} & 71.02\\
LAR-large w/o EI & 74.49 & 70.66 & 76.29 & 70.0 & 63.56 & 71.0\\
LAR-large & \textbf{79.45} & 67.14 & \underline{77.22} & \textbf{72.43} & \textbf{68.48} & \textbf{72.94}\\
\bottomrule
\end{tabular}
}
\caption{\label{table-supervised}
The experimental results on in-domain supervised setting. EI denotes enhanced inference with a general language model.
}
\end{table*}

\renewcommand{\thefootnote}{1}

\section{Experiment}
\subsection{Experimental Settings}
\textbf{Dataset.} Following previous related works, we conduct our experiments on these classic NER datasets: CoNLL03 \cite{CoNLL03-2003}, CrossNER \cite{CrossNER-2020}, MIT-Movie \cite{MIT-Movie-2013} and MIT-Restaurant \cite{MIT-Restaurant-2013}. CoNLL03, including four generic entity types: PER, LOC, ORG and MISC, generally is chosen as the source domain dataset. CrossNER, which is commonly used as the target domain dataset, is drawn from Wikipedia and contains five different domain datasets: politics, natural science, music, literature, and AI. Movie and Restaurant corpus consist of user utterances for movie and restaurant domains with 12 and 8 classes. For all datasets, we follow their official splits of train, validation, and test sets. The statistics of these datasets are summarized in Table \ref{table-statistics}.\\
\\
\textbf{Configuration.} In our experiments, we use Flan-T5 \cite{Flan-T5-2022} as the fine-tuned foundation language model and GPT-3.5-turbo as external knowledge access assistant. For the enhanced inference technique, we only provide one example as demonstration on both supervised and zero-shot out-of-domain setting, which aims to offer LLM a canonical output format rather than context knowledge. We use AdamW \cite{AdamW-2017} as optimizer with learning rate=5e-5 and batch size=8 for all training scripts. The code of this work is implemented through PyTorch, Transformers \cite{Transformers-2020} and OpenAI API \footnote{https://platform.openai.com/docs/overview}.

\begin{table*}
\centering
\resizebox{\textwidth}{!}{
\begin{tabular}{lccccccc}
\toprule
\textbf{Model} & \textbf{Additional Annotation} & \textbf{Politics} & \textbf{Science} & \textbf{Music} & \textbf{Literature} & \textbf{AI} & \textbf{Average}\\
\midrule
FactMix-base \cite{FactMix-2022} & No & 44.66 & 34.13 & 23.75 & 28.89 & 32.09 & 32.70\\
PLTR-base \cite{PLTR-2023} & No & 47.56 & 34.87 & 30.52 & 30.80 & 33.87 & 35.52\\
UniNER-7B \cite{UniversalNER-2023} & Yes & 60.80 & \underline{61.10} & \underline{65.0} & 59.40 & \underline{53.50} & 59.96\\
UniNER-13B \cite{UniversalNER-2023} & Yes & \underline{61.40} & \textbf{63.50} & \underline{64.50} & \underline{60.90} & \textbf{54.20} & \underline{60.9}\\
\midrule
LAR-base & No & \underline{68.56} & 45.54 & 61.15 & \underline{61.99} & 43.07 & 56.06\\
LAR-base+ & Yes & \textbf{74.33} & 52.18 & \textbf{67.07} & \textbf{64.38} & 49.98 & \textbf{61.58}\\
\bottomrule
\end{tabular}
}
\caption{\label{table-ood}
The experimental results on zero-shot out-of-domain setting. For LAR-base+, 20\% samples randomly selected from each target domain dataset are labeled by prompting ChatGPT.}
\end{table*}

\subsection{Baselines}
To explore the performance of our method, we compare it to some most advanced methods for cross-domain NER, which include both supervised and  zero-shot out-of-domain settings.\\
$\bullet$ \textbf{FLAIR} \cite{FLAIR-2018}: This leverages the internal states of a character language model to produce contextual string embedding and then integrate them into the NER model.
\\
$\bullet$ \textbf{Cross-domain LM} \cite{CrossDomainLM-2019}: This employs a parameter generation network to combine cross-domain language modeling and NER, thereby enhancing The model performance.
\\
$\bullet$ \textbf{DAPT} \cite{DAPT-2021}: DAPT employs a large unlabeled corpus related to a specific domain, which is based on BERT and fine-tunes it on the CrossNER benchmark.
\\
$\bullet$ \textbf{LANER} \cite{LANER-2022}: This method introduces a new approach for cross-domain named entity recognition by utilizing an autoregressive framework to strengthen the connection between labels and tokens.
\\
$\bullet$ \textbf{LightNER} \cite{LightNER-2021}: This work utilizes a pluggable prompting method to improve NER performance in low-resource settings.
\\
$\bullet$ \textbf{FactMix} \cite{FactMix-2022}: A two-step rationale-centric data augmentation method to improve the model’s generalization ability, demonstrating strong few-shot learning ability on cross-domain NER task.
\\
$\bullet$ \textbf{CP-NER} \cite{Cp-NER-2023}: This approach utilizes textto-text generation grounding domain-related instructors to transfer knowledge to new domain NER tasks, achieving state of the art performance.
\\
$\bullet$ \textbf{PLTR} \cite{PLTR-2023}: A prompt learning method with type-related features to address implicit knowledge transfer issue in a limited data scenario, getting remarkable performances under both the supervised and zero-shot out-of-domain settings.
\\
$\bullet$ \textbf{UniNER} \cite{UniversalNER-2023}: This study explores targeted distillation with mission-focused instruction tuning to train student models that can excel in a
broad application, also outperforming general large language models by a large margin.

\subsection{Results on supervised OOD setting}
We evaluate the low-resource supervised learning ability of our method on CrossNER datasets and the results are shown in Table \ref{table-supervised}. The F1 scores indicate that our method can compete with the most advanced cross-domain NER method. Especially while the base model is Flan-T5-large, our method achieves the highest average score on five target domain datasets. As well as, we report the performances of two variants which are not enhanced by GPT-3.5. They outperform all baselines other than the recent work Cp-NER, but they still perform 1.5 percent less than the enhanced models on average. It is clear that enhanced inference by using GPT-3.5 in the label reassignment process generally improves model performance not requiring any extra context knowledge but it struggles in certain specific domains such as science. Without enhanced inference, our method based on adaptive label alignment also outperforms most of the baselines.

\subsection{Results on zero-shot OOD setting}
The results of our method and baselines on the zero-shot out-of-domain setting are reported in Table \ref{table-ood}. LAR-base is trained on the source domain dataset and directly performs on downstream datasets, while LAR-base+ represents label augmentation by using GPT-3.5, which is prompted to randomly produce predicted labels for 20\% samples on each target domain datasets. It's clear that our method achieves excellent performances compared to the baselines, getting 56.06\% and 61.58\% F1 scores respectively. Even without label augmentation, our method also approaches the most advanced open NER method. It only underperforms UniNER-7B by 3.9 percentage points. These experimental results indicate our method has strong out-of-domain generalization ability if the source data distribution is partially close to that of target domains.

\begin{table}
\centering
\resizebox{0.48\textwidth}{!}{
\begin{tabular}{lcccc}
\toprule
\textbf{Dataset} & \textbf{Comparable baseline} & \textbf{LAR} & \textbf{w/ EI}\\
\midrule
CoNLL03 & \textbf{92.93} \cite{LightNER-2021} & 92.63 & -\\
ACE05 & 86.69 \cite{UniversalNER-2023} & 85.56 & \textbf{86.79}\\
AI & 56.36 \cite{DAPT-2021} & 56.80 & \textbf{58.27}\\
Music & 73.39 \cite{DAPT-2021} & 73.33 & \textbf{74.69}\\
GENIA & \textbf{77.54} \cite{UniversalNER-2023} & 77.08 & 77.21\\
\bottomrule
\end{tabular}
}
\caption{\label{table-ID}
The results on supervised in-domain setting. We use Flan-T5-large as the fine-tuned foundation model.
}
\end{table}

\subsection{Results on supervised in-domain setting}
To validate the stability of our approach, we also evaluate the model on the supervised in-domain setting. In this case, all entity types are visible, so the model doesn't entail the label alignment process. However, enhanced inference still can be applied for label reassignment. For example, if we observe the model performs poorly when classifying entity types, the inference process may be enhanced with the help of ChatGPT. Specifically, the supervised in-domain experiment results of our method are demonstrated in Table \ref{table-ID}. On the CoNLL03 dataset, LAR without enhanced inference achieves close performance with the baseline. On the ACE05-Ent dataset, the entities with type "GPE" are reassigned by GPT-3.5, with the performance improved by around 1.2\%. On the AI dataset and the Music dataset, our method significantly exceeds the baseline method through enhancing inference on type "programming language", "country", "instrument", "conference", and "person". We also use enhanced inference on type "DNA" of GENIA dataset, which leads to a little improvement on the performance. Even though our method is not designed for in-domain NER, it still exhibits performance close to some general NER methods.

\begin{table}
\centering
\resizebox{0.48\textwidth}{!}{
\begin{tabular}{l|c|cc|cc}
\toprule
\multirow{2}*{\textbf{Dataset}} & \multirow{2}*{\textbf{Proportion}} & \multicolumn{2}{c}{\textbf{Base}} & \multicolumn{2}{c}{\textbf{Large}}\\
 & & Ori & LA & Ori & LA\\
\midrule
\multirow{3}*{\textbf{Movie}} & 1\% & 50.14 & 52.84 & 56.73 & 54.87\\
 & 5\% & 60.24 & 62.69 & 62.63 & 64.41\\
 & 10\% & 64.50 & 65.56 & 66.38 & 65.99\\
\midrule
\multirow{3}*{\textbf{Restaurant}} & 1\% & 55.89 & 58.78 & 60.67 & 61.50\\
 & 5\% & 69.16 & 70.37 & 71.29 & 71.77\\
 & 10\% & 72.83 & 74.15 & 74.11 & 75.08\\
\bottomrule
\end{tabular}
}
\caption{\label{table-LA}
The results of ablation study on label alignment. Ori denotes direct transfer learning from the source domain to the target domain without using label alignment.
}
\end{table}

\section{Analysis}
\subsection{Label Alignment}

\begin{table}
\centering
\resizebox{0.48\textwidth}{!}{
\begin{tabular}{l|c|cc}
\toprule
\textbf{Dataset} & \textbf{Type} & \textbf{Direct Inference} & \textbf{Enhanced Inference}\\
\midrule
\multirow{5}*{\textbf{Politics}} & politician & 79.59 & 85.36 (+5.77)\\
 & person & 77.97 & 84.75 (+4.78)\\
 & country & 89.95 & 95.93 (+5.98)\\
 & event & 79.59 & 90.26 (+10.67)\\
 & election & 94.01 & 95.62 (+1.61)\\
\midrule
\multirow{5}*{\textbf{Science}} & scientist & 84.93 & 92.57 (+7.64)\\
 & award & 82.19 & 90.41 (+8.22)\\
 & protein & 83.33 & 82.05 (-1.28)\\
 & university & 86.92 & 85.38 (-1.54)\\
 & miscellaneous & 84.62 & 70.77 (-13.85)\\
\midrule
\multirow{5}*{\textbf{Music}} & band & 92.64 & 93.72 (+1.08)\\
 & song & 89.87 & 94.71 (+4.84)\\
 & album & 91.87 & 94.58 (+2.71)\\
 & genre & 92.40 & 88.91 (-3.49)\\
 & instrument & 77.27 & 95.45 (+18.18)\\
\bottomrule
\end{tabular}
}
\caption{\label{table-LR}
The results of type-wise quantitative analysis about direct inference (by Flan-T5-large) and enhanced inference (by GPT-3.5). 
}
\end{table}

To explore the functionality and effectiveness of label alignment, we conducted an ablation experiment on two datasets under the low-resource setting, and the results are shown in Table \ref{table-LA}. We set CoNLL03 as the source domain dataset and MIT-Movie, MIT-Restaurant as the target domain datasets. To imitate the low-resource supervised learning scenario, we randomly assigned 1\%, 5\%, and 10\% of the training set as supervised data on the target domains. For every result of the original transfer and the label alignment transfer, we report the average scores of 5 experiments. From the results, the model performance with label alignment is clearly higher than that of original transfer learning which directly learns entity knowledge from different domains regardless of label relationships. With less supervised data and smaller model setups, this gain effect becomes even more pronounced. It indicates that label alignment is able to augment model transfer ability in low-resource supervised learning scenarios.

\begin{figure}
    \includegraphics[width=0.5\textwidth, right]{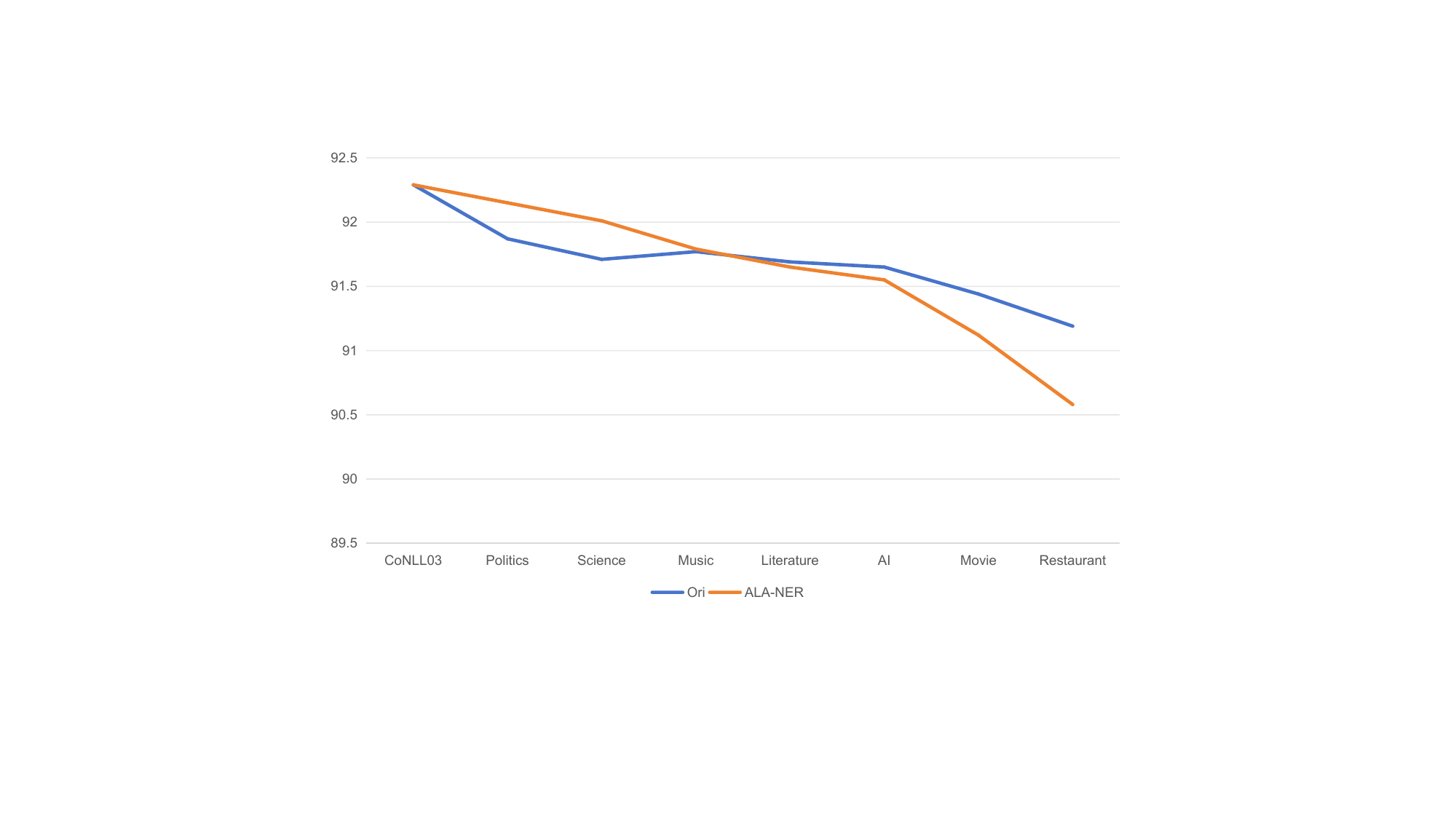}
    \caption{The trend of performance degradation of our model when it is applied to continuous learning.}
    \label{fig:cl-performance}
\end{figure}

\begin{figure*}[t]
    \includegraphics[width=\textwidth, right]{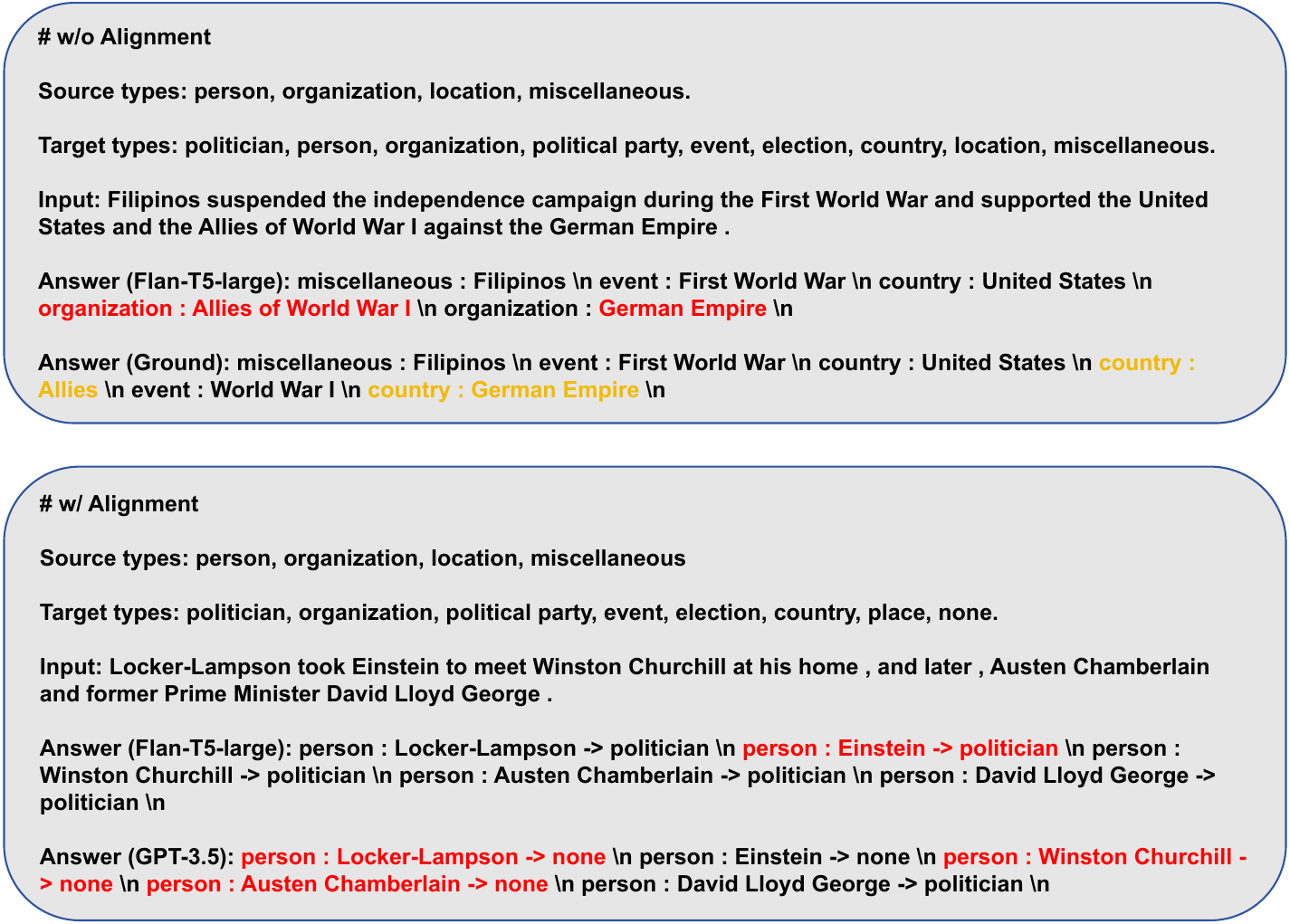}
    \caption{The successful and failure extraction cases of our method with and without label alignment. The instances that are predicted incorrectly are marked in red and the missing instances are marked in orange. The samples and ground labels originate from the test set of the Politics dataset.}
    \label{fig:case}
\end{figure*}

\subsection{Label Reassignment}
Label reassignment is an important part of our method, which greatly influences the model performance. There are two inference forms in the label reassignment process, direct inference and enhanced inference. The former depends on knowledge and extrapolation of the model itself, while the latter resorts to an extra knowledge source to enhance model ability for inference. In Table \ref{table-LR}, we investigate the accuracy of direct inference and enhanced inference on target domain types. From the results, enhanced inference demonstrates higher scores than direct inference on most types, especially event (+10.67\%), award (+8.22\%) and instrument (+18.18\%). However, enhanced inference also performs badly on some types, e.g. miscellaneous (-13.85\%) and genre (-3.49\%). As well as, using ChatGPT helps discriminate similar types when target domain types conflict with source domain types e.g. politician and person. Because we do not provide in-context knowledge for ChatGPT, the results of enhanced inference also apply to the zero-shot out-of-domain setting. In addition, we observe GPT-3.5 is more confident in type classification than Flan-T5-large through repeated sampling from answers and logits when candidate entity types are similar, which indicates enhanced inference is capable of handling more difficult samples.

\subsection{Continual Learning Capability}
Many cross-domain NER methods suffer from the catastrophic amnesia problem and are not able to perform in the continual learning paradigm. Figure \ref{fig:cl-performance} shows the performance of our method trained by continual learning on a set of NER datasets. Initially, the Flan-T5-base is trained on CoNLL03, getting 92.29 scores. We then add the training data from target domains in turn and continuously observe its performance on the CoNLL03 dataset. The results show that there is a decreasing trend regardless of whether label alignment is used. But as more target domain data is used for continuous learning, the LAR decline is significantly more gradual than the original training style. One possible reason is that our approach eliminates the problem of overlapping labels between different data sets, avoiding knowledge conflicts to a certain extent.

\subsection{Case Analysis}
In this section, we display two extraction cases presented in Figure \ref{fig:case} to show the predictive preferences of our model. Firstly, the top part of the picture is an extraction case of our model without label alignment. There are some overlapped entity types between the source type collection and the target type collection, which results in the output of the model being more biased towards the source domain knowledge that has a more significant impact on the model than the target domain knowledge. For this example, the model tends to incorrectly classify the countries as organizations, which can be rectified and restricted according to the type relationship mapping after label alignment. On the other hand, the bottom of the picture shows a sample predicted by Flan-T5-large and GPT-3.5 based on label alignment. Clearly, GPT-3.5 discriminates between entity types about person and politician more accurately. In this case, there is no obvious evidence in the context to indicate that Locker-Lampson, Einstein, and Winston Churchill are politicians. This result implies that GPT-3.5 is more cautious in making inferences of entity type than the fine-tuned language model which is limited with scarce attainable supervised data.

\section{Conclusion}
In this study, we introduce a label alignment and reassignment approach to address the label conflict problem on the cross-domain NER task. Our method incorporates a generalist large language model to enhance entity type inference in prediction, which enables our model to perform well in both the supervised and zero-shot out-of-domain setting. To validate the effectiveness of our method, we conduct a series of empirical experiments across different settings and in-depth analysis for each important part of the approach. The experimental results demonstrate our model has the good performances on out-of-domain NER datasets and the robustness on the in-domain and continual learning settings.

\bibliography{anthology,custom}
\bibliographystyle{acl_natbib}

\end{document}